\def\year{2019}\relax
\documentclass[letterpaper]{article} 

\usepackage{aaai19}  
\usepackage{times}  
\usepackage{helvet}  
\usepackage{courier}  
\usepackage{url}  
\usepackage{graphicx}  
\usepackage{url}            
\usepackage{booktabs}       
\usepackage{amsfonts}       
\usepackage{nicefrac}       
\usepackage{microtype}      
\usepackage{algorithmic}
\usepackage{algorithm}
\usepackage{wrapfig}
\usepackage{lipsum}
\usepackage{xspace}
\usepackage{graphicx} 
\usepackage{subfigure} 
\usepackage{caption}
\usepackage{paralist}
\usepackage{float}
\usepackage{tabularx}
\usepackage{multirow}
\usepackage{diagbox}
\usepackage{tikz}
\usepackage{booktabs}
\usepackage{array, makecell} 
\usepackage{natbib}
\usepackage{amsmath,amssymb}
\usepackage[inline]{enumitem}
\usepackage{acronym}

\title{Dialogue Generation: From Imitation Learning to\\ Inverse Reinforcement Learning}
\author{
Ziming Li\textsuperscript{1} \ and
Julia Kiseleva\textsuperscript{1,}\textsuperscript{2} \ and
Maarten de Rijke\textsuperscript{1}\\
\textsuperscript{1}University of Amsterdam\\
\textsuperscript{2}Microsoft Research AI\\
\{z.li, derijke\}@uva.nl,
julia.kiseleva@microsoft.com
}

\DeclareMathOperator{\E}{\mathbb{E}}

\acrodef{RL}{Reinforcement Learning}
\acrodef{DRL}{Deep Reinforcement Learning}
\acrodef{IRL}{Inverse Reinforcement Learning}
\acrodef{SERP}{search engine result page}
\acrodef{IR}{Information Retrieval}
\acrodef{MDP}{Markov Decision Process}
\acrodef{MaxEnt-IRL}{Maximum Entropy Inverse Reinforcement Learning}
\acrodef{DM-IRL}{Distance Minimization Inverse Reinforcement Learning}
\acrodef{MMI}{Maximum Mutual Information} 
\acrodef{DNN}{Deep Neural Networks}
\acrodef{RNN}{Recurrent Neural Networks}

\begin{document}
\date{}

\maketitle

\begin{abstract}
The performance of adversarial dialogue generation models relies on the quality of the reward signal produced by the discriminator. The reward signal from a poor discriminator can be very sparse and unstable, which may lead the generator to fall into a local optimum or to produce nonsense replies. To alleviate the first problem, we first extend a recently proposed adversarial dialogue generation method to an adversarial imitation learning solution. Then, in the framework of adversarial inverse reinforcement learning, we propose a new reward model for dialogue generation that can provide a more accurate and precise reward signal for generator training. We evaluate the performance of the resulting model with automatic metrics and human evaluations in two annotation settings. Our experimental results demonstrate that our model can generate more high-quality responses and achieve higher overall performance than the state-of-the-art. 
\end{abstract}

\section{Introduction}
\label{sec:introduction}
The task of an open-domain dialogue system is to generate sensible dialogue responses given a dialogue context \citep{ritter2010unsupervised,shang2015neural,li2016deeprl,xing2017topic}. 
There are two broad directions for training a dialogue generation system: the first employs defined rules or templates to construct possible responses and the second builds a chatbot to learn the response generation model with a machine translation framework from social dialogue collections~\citep{shang2015neural,sordoni2015neural, serban2016hred,serban2017vhred}.
Sequence-to-sequence (Seq2Seq) models enjoy the advantages of scalability and language independence and the maximum likelihood estimation objectives make it simple to train them. 
However, in dialogue generation, the trained model suffers from generating dull and generic responses such as ``I don't know"~\citep{sordoni2015neural, serban2016hred, li2016diversity, li2017adversarial}, which are meaningless. \citet{li2016diversity} suggest that ``by optimizing for the likelihood of outputs given inputs, neural models assign a high probability to `safe' responses." 
To alleviate this problem, \citet{li2016deeprl} introduced a neural \ac{RL} generation method to generate coherent and interesting dialogues by optimizing the manually defined reward function covering ideal dialogue properties.
However, a handcrafted reward function is expensive to maintain and does not generalize over different domains \citep{guided_cost_learning,fu2017learning}. 
Especially for open-domain dialogue systems, it is hard to decide what knowledge is essential to design a proper reward function \citep{li2017adversarial}. 
Additionally, the accuracy of defined reward functions can degrade when the dialogue context becomes more complex. \citet{li2017adversarial} use adversarial training for dialogue generation where they jointly train two systems, a generative model to produce response sequences and a discriminator to distinguish between the human-generated dialogues and the machine-generated ones. 
Feedback from the discriminator is used as a reward to push the generator to produce more realistic replies. 
The discriminator takes a dialogue consisting of a context-reply pair as input and outputs the probability that this dialogue is from real human dialogues. 

In \citet{li2017adversarial}, during generator training, the reward of each generated word during decoding should be supplied and Monte Carlo search is applied to estimate the reward for each word position. 
A potential problem is that the returned reward from the discriminator could be very sparse and unstable, which may lead the generator to produce unintended and nonsense replies. 
Moreover, \citet{li2017adversarial} put no constraints on the generator policy, which can result in two problems. 
First, the learned policy may prefer to generate general responses. 
Second, the training step can easily get stuck in a local optimum, which leads the generator to produce identical responses regardless of the input context or even worse -- the outputs from the generator are always the same ungrammatical sentence. 

In this paper, we first extend the adversarial dialogue generation method introduced by \citet{li2017adversarial} to a new model, DG-AIL, which incorporates an entropy regularization term to the generation objective function. 
This addition can alleviate the problem of model collapse. 
Then we adopt adversarial inverse reinforcement learning to train a dialogue generation model, DG-AIRL.
This method enables us to both make use of an efficient adversarial formulation and recover a more precise reward function for open-domain dialogue training. 
Unlike~\citet{shi2018towards}, we design a specific reward function structure to measure the reward of each word in generated sentences while taking account of the dialogue context. 
We also consider two human evaluation settings to assess the overall performance of our model. 

To summarize, we make the following contributions:
\begin{itemize}[nosep]
\item A novel reward model architecture to evaluate the reward of each word in a dialog, which enables us to have more accurate signal for adversarial dialogue training;
\item A novel Seq2Seq model, DG-AIRL, for addressing the task of dialogue generation built on adversarial inverse reinforcement learning;
\item An improvement of the training stability of adversarial training by employing causal entropy regularization; 
\end{itemize}

\section{Background}
\label{sec:background}

\subsubsection{Preliminaries.}
We build our dialogue system as a Markov Decision Process (MDP), which is defined by a tuple $(S, A, \tau, r, \gamma)$, where $S$ and $A$ are the state space and action space, respectively, $\tau$ is the transition probability, and  $\tau(s, a, s')$ is the probability of transitioning from state $s$ to state $s'$ under action $a$ at time $t$:
\begin{equation}
     \tau (s' \mid s, a) = P(s_{t+1}=s' \mid s_t = s, a_t=a).
\end{equation}
Here, $r(s,a)$  is the immediate reward after taking action $a$ in state $s$; $\gamma \in [0,1]$ is a discount factor.

The \emph{dialogue response strategy} is represented by a policy, which is a mapping $\pi \in \Pi$ from states $s \in S$ and actions $a \in A$ to $\pi(a|s)$, which is the probability of performing action $a_t=a$ by the user when in state $s_t=s$:
\begin{equation}
  \mathbb \pi (a|s) =  P(a_t=a\mid s_t=s).
\end{equation}

\subsubsection{Maximum causal entropy.}
Motivated by the task of decision prediction in sequential interactions, \citet{ziebart2010causal-entropy} propose to use maximum causal entropy to model the availability and influence of sequentially revealed side information. 
The causal entropy of policy $\pi$ is defined as:
\begin{equation}
H(\pi) \triangleq E_{\pi} [- \log \pi (a | s )]),
\label{eq:causal-ent}
\end{equation}
which measures the uncertainty presented in policy $\pi$ \citep{ziebart2010causal-entropy}.

\subsubsection{Maximum entropy inverse reinforcement learning (\acs{MaxEnt-IRL}).}
\label{sec:max-irl}
Given a set of demonstrated (expert) behavior, which can be seen as the trajectories resulting from executing expert policy $\pi_E$, \ac{IRL} aims to find a reward function that can rationalize the given behavior. 
In \ac{MaxEnt-IRL}~\citep{ziebart_aaai_2008}, the demonstrated behavior $D_{demo} = \{\zeta_1, \ldots, \zeta_{N}\}$ is assumed to be the result of an expert acting stochastically and near-optimally with respect to an unknown reward function.
Trajectories with equivalent rewards have equal probability to be selected and trajectories are sampled from the distribution: 
\begin{equation}
\mbox{}\hspace*{-2mm}
p(\zeta_i\mid \theta)\! =\! \frac{1}{Z(\theta)} \exp^{r_\theta(\zeta_i)}\! =\! \frac{1}{Z(\theta)} \exp^{\sum_{t=0}^{|\zeta_i|-1}r_\theta(s_t, a_t)},%
\hspace*{-1.5mm}\mbox{}
\label{eq:max-irl}
\end{equation}
where $Z(\theta) = \int \exp(r_\theta (\zeta)) d \zeta$ is the partition function and $r_\theta$ is the reward function, which takes a state-action pair as input. 
\ac{MaxEnt-IRL} maximizes the likelihood of the demonstrated data $D_{demo}$ under the maximum entropy (exponential family) distribution and the objective is given as:
\begin{equation}
L(\theta) = -\E_{\zeta \sim D_{demo}} r_\theta(\zeta) + \log Z.
\label{eq:max-likelihood-obj}
\end{equation}
This task can be seen as a classification problem where each trajectory represents one class. 
However, it is difficult to apply vanilla \ac{MaxEnt-IRL} to complex and high-dimensional settings since computing the partition function $Z(\theta)$ is intractable in the original method.
To overcome this drawback, \citet{guided_cost_learning} combine sample-based maximum entropy IRL with forward reinforcement learning to estimate the partition function $Z$, where: 
\begin{equation}
\mbox{}\hspace*{-1.75mm}
L(\theta) = -\E_{\zeta_i \sim p} r_\theta(\zeta_i) + \log \left(\E_{\zeta_j \sim q} \left[\frac{\exp(r_\theta(\zeta_j))}{q(\zeta_j)}\right]\right)\!\!.
\hspace*{-1mm}
\label{eq:guided-cost-obj}
\end{equation}
Here, $p$ represents the distribution of demonstrated samples, while $q$ is the background distribution for estimating the partition function $\int \exp(r_\theta (\zeta)) d \zeta$. 
This work alternates between updating the reward function $r_\theta$ to maximize the likelihood of the demonstrated data and optimizing the background distribution $q$ to minimize the variance of the importance sampling estimation.

\subsubsection{Generative adversarial imitation learning.}
Recovering the true reward function is intractable in real scenarios \citep{ziebart_aaai_2008,gan_imitation,fu2017learning}.
In previous research, if only the optimal policy is pursued, imitation learning is used to rebuild the policy network directly by skipping recovering reward functions. 
\citet{gan_imitation} cast the problem of \ac{IRL} as an optimization problem in the paradigm of Generative Adversarial Networks (GANs), where the discriminator corresponds to the reward function and the generator corresponds to the policy used to sample trajectories. 
The optimization problem is given as:
\begin{equation}
\max_{r \in R} \left(\min_{\pi \in \Pi} -\lambda H(\pi) - \mathbb{E}_{\pi}[r(s,a)]\right) + \mathbb{E}_{\pi_E}[r(s,a)].
\label{eq:opt-max-irl}
\end{equation}
The optimization of Eq.~\ref{eq:opt-max-irl} is converted to an imitation learning algorithm:
 \begin{equation}
 \min_{\pi \in \Pi} -\lambda H(\pi) + D_{JS}(\rho_\pi, \rho_{\pi_E}),
\label{eq:ail}
\end{equation}
which finds a policy $\pi$ whose occupancy measure $\rho_\pi$ minimizes the Jensen-Shannon divergence to the expert's policy $\pi_E$ (the policy of demonstrated data). 
The occupancy measure $\rho_\pi$ can be interpreted as the unnormalized distribution of state-action pairs that an agent encounters when navigating the environment with policy $\pi$. 
Eq. \ref{eq:ail} can be solved by finding a saddle point $(\pi, D)$ of the expression
\begin{equation}
\begin{split} 
 &\mathbb{E}_{\pi}[-\log(D(s,a))]+ {}\\
 &\quad \mathbb{E}_{\pi_E}[-\log(1 - D(s,a))] - \lambda H(\pi),
\end{split}
\label{eq:ail-gan}
\end{equation}
where $D$ is a binary classifier to distinguish state-action pairs of $\pi$ and $\pi_E$.

\section{Method}
\label{sec:method}
In this section, we will first extend the work by~\citet{li2017adversarial} to the framework of adversarial imitation learning, and then introduce our main model, which applies adversarial inverse reinforcement learning to train a dialogue system.
\subsection{Problem setting}
In a dialogue setting, the word sequence $\langle w_1, w_2, \ldots, w_t\rangle$ in an utterance can be regarded as corresponding actions $\langle a_1, a_2, \ldots, a_n\rangle$ taken by the policy network at different time steps. 
We use a state function $f$ to compress the dialogue context and the words already generated in the current utterance to an intermediate representation, which will be regarded as the current state. 
For example, $s_0= f(p)$ represents the state at time step 0 and it takes the dialogue context $p$ as input. 
State $s_t$ is given as $s_t = f(p, a_1, a_2, \ldots, a_{t-1})$. 
In this work, we limit the range of the dialogue context to the utterances in the last two conversation turns. 

Given an initial state $s_0$ representing the history of previous dialogues, a well-trained dialogue system should reply with a reasonable sentence $\langle w_0, w_1, \ldots, w_t\rangle$ generated by selecting a specific word at different time steps. 
The length $t$ is automatically decided by the policy network. 
We aim to find the optimal policy $\pi(a_t|s_t)$ that selects the most appropriate word at each time step.

\subsection{Dialogue generation with adversarial imitation learning (DG-AIL)}
In the framework of adversarial imitation learning, we aim to train a dialogue system to imitate the way humans talk by observing real human dialogues. 
This model DG-AIL can be regarded as an extension of the work of~\citet{li2017adversarial}. 

Unlike previous work, we do not only consider the difference between the distributions of real dialogues and generated dialogues but also take into account how the previous state-action pairs affect future words under a specific policy network $\pi$, which can be measured by the causal entropy $H(\pi)$.  

\begin{figure}[ht]
\centering
   \includegraphics[clip, width=0.95\columnwidth]{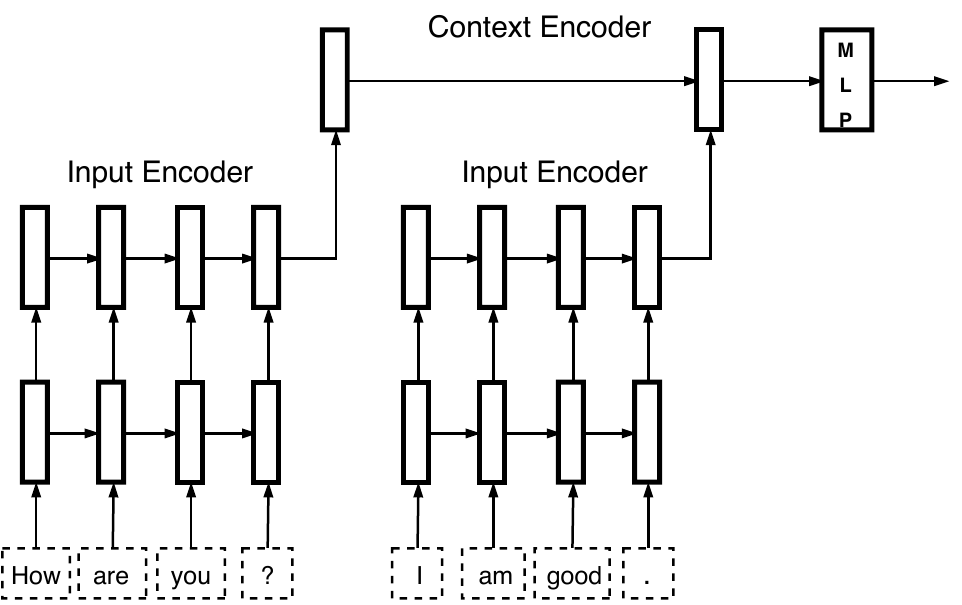}
   \caption{Discriminator architecture in DG-AIL.}
   \label{fig:disc}
   \vspace*{-0.5\baselineskip}
\end{figure}

In adversarial learning, the task of the discriminator $D$ is to distinguish dialogues from the true data distribution and dialogues from the generator.
As shown in Fig.\ref{fig:disc}, we adopt a hierarchical structure to represent the discriminator model. 
The first layer is an input encoder that compresses the utterances from each speaker in the conversation. 
Then, a context encoder sequentially takes as input the utterance representations and generates a final state to represent the whole dialogue. 
In the end, the final state is fed to a binary classifier that predicts whether the dialogue is real or fake with a confidence value. 

According to Eq. \ref{eq:ail-gan}, the gradient of the discriminator parameters is given as:
\begin{equation}
\begin{split}
\label{eq:disc-grad}
 \nabla D_\theta = {}&\mathbb{E}_{\zeta \sim \pi}[\nabla_\theta \log(D_\theta(s, a))] + {}\\
 & \quad \mathbb{E}_{\zeta \sim \pi_E}[\nabla_\theta \log(1 - D_\theta(s, a))].
 \end{split}
\end{equation}
The generative model $G$ attempts to generate high-quality human-like responses to confuse the discriminative classifier $D$ while maintaining a high policy entropy. 
The gradient to update the generator parameters can be inferred from Eq.~\ref{eq:ail-gan} as follows:
\begin{equation}
\begin{split}
\mbox{}\hspace*{-1mm}
\nabla G_\phi = {}& \nabla_\phi [- \lambda H(\pi_\phi) - \mathbb{E}_{\zeta \sim \pi_\phi}[D_\theta(s, a)] ]\\
{} = {} & -  \mathbb{E}_{\zeta \sim \pi_\phi} \nabla_\phi [\log(\pi_\phi (a|s))] (Q_\theta(s, a) - {}\\
& \phantom{-\mathbb{E}_{\zeta \sim \pi_\phi} \nabla_\phi [\log(\pi_\phi (a|s))] (Q}
\lambda  \log \pi_\phi (a|s)),
\end{split}
\label{eq:generator-grad-ail}
\end{equation}
where $Q_\theta(s, a) = \mathbb{E}_\zeta[\log (D_\theta (s, a)) | s_0 = \bar{s}, a_0 = \bar{a}]$ is estimated with Monte Carlo search.

\subsection{Dialogue reward learning with adversarial inverse reinforcement learning (DG-AIRL)}
Our main model DG-AIRL adopts inverse reinforcement learning techniques to train a dialogue generation model. 
We assume that human participants in a dialogue are using a true reward function that guides them to formulate a policy to react with different replies to different dialogue contexts. 
Unlike the use of a classifier to supply a reward signal in the model DG-AIL, the reward model in DG-AIRL has a more specific architecture to evaluate the reward for each state-action pair, which can provide more accurate and precise reward signal to update the generator.   
\subsubsection{Dialogue response policy.}
In maximum entropy inverse reinforcement learning, the reward model (discriminator) attempts to assign high rewards to demonstrated trajectories (from the expert policy) and low rewards to sampled trajectories from other policies. 
In this way, when the reward function is fixed, the expert policy can be found by solving a common reinforcement learning problem:
\begin{equation}
G_\phi(r_\theta) = \arg\min_{\pi \in \Pi}- \lambda H(\pi) - \mathbb{E}_{\zeta \sim \pi}[r_\theta(\zeta)],
\label{eq:generator-obj}
\end{equation}
where $\zeta$ represents the sampled dialogues and $H(\pi)$ is the causal entropy regularization term; 
$r_\theta (\zeta)$ is the reward of dialogue $\zeta$ that can be accessed from the reward model. 
The goal of the generator is to generate dialogues that can achieve higher rewards from the reward model. 
The found policy maximizes the expected cumulative reward while maintaining high-entropy. 

The derivative can be inferred as follows:
\begin{equation}
\begin{split}
\nabla_\phi G(r)_\phi  = {} & \nabla_\phi [- \lambda H(\pi_\phi) - \mathbb{E}_{\zeta \sim \pi_\phi}[r_\theta(\zeta)] ]\\
{} = {} & -  \mathbb{E}_{\zeta \sim \pi_\phi} \nabla_\phi [\log(\pi_\phi (\zeta))] (r_\theta(\zeta) - {}\\
& \phantom{-  \mathbb{E}_{\zeta \sim \pi_\phi} \nabla_\phi [\log(\pi_\phi (\zeta))] (r} \lambda  \log \pi_\phi (\zeta)).
\end{split}
\label{eq:generator-grad-airl}
\end{equation}
If we decompose dialogue $\zeta$ into different time steps, the gradient is given as:
\begin{equation}
\begin{split}
\mbox{}\hspace*{-1mm}
\nabla_\phi G(r)_\phi  = &  -  \mathbb{E}_{\zeta \sim \pi_\phi} \nabla_\phi [\log(\pi_\phi (\zeta))] (r_\theta(\zeta) -{} \\
& \phantom{-  \mathbb{E}_{\zeta \sim \pi_\phi} \nabla_\phi [\log(\pi_\phi (\zeta))] (r}
\lambda \log \pi_\phi (\zeta)) \\
= & \textstyle
 - \sum_t \mathbb{E}_{\pi_\phi (a_t\mid s_t )} \nabla_\phi [\log(\pi_\phi (a_t\mid s_t))] \\
 & \textstyle
 \mbox{}\hspace*{1.5cm}(r_\theta(\zeta_{t:T}) - \lambda \log \pi_\phi (a_t\mid s_t)).
\end{split}
\label{eq:generator-grad-time}
\end{equation}

The reward $r_\theta(\zeta_{t:T})$ of a partial dialogue from time $t$ to $T$ is estimated with Monte Carlo search.
\begin{figure}[t]
\centering
   \includegraphics[clip, width=0.95\columnwidth]{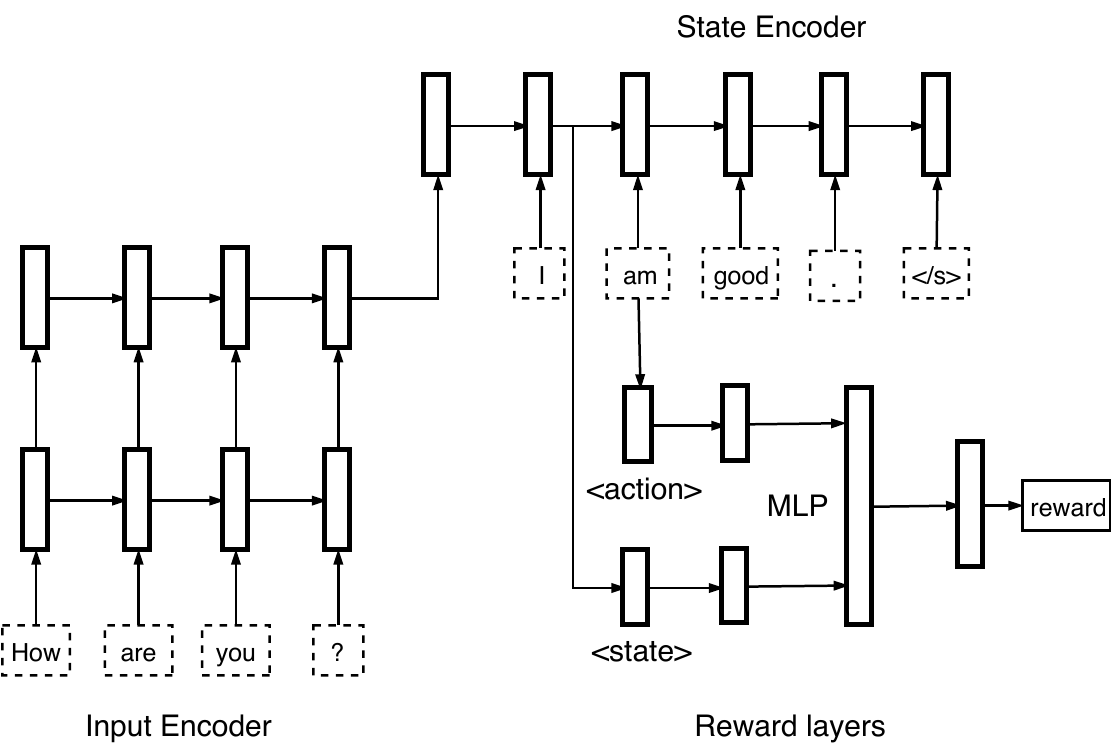}
   \caption{Reward model architecture in DG-AIRL.}
   \label{fig:reward_model}
   \vspace*{-0.5\baselineskip}   
\end{figure}

\subsubsection{Reward learning.}
Following prior work on sample-based maximum entropy IRL (Eq.~\ref{eq:guided-cost-obj}), the objective (loss function) of our reward model is given as:
\begin{equation}
\mbox{}\hspace*{-1.5mm}
L(\theta) \!=\! -\mathbb{E}_{\zeta_i \sim \pi_E}r_\theta(\zeta_i) + \log \mathbb{E}_{\zeta_j \sim \pi} \left(\frac{\exp(r_\theta(\zeta_j))}{q(\zeta_j)}\right)\!,
\hspace*{-1.5mm}\mbox{}
\label{eq:reward-obj}
\end{equation}
where $\pi_E$ denotes the policy of demonstrated trajectories and $\pi$ the policy of background samples. 
The term $q$ denotes the background distribution from which dialogues $\zeta_j$ were sampled. 
In our setting, $q$ is the distribution of dialogues generated with the current dialogue policy $\pi$. 
We use $D_{demo}$ and $D_{samp}$ to represent the set of dialogues generated with policy $\pi_E$ and $\pi$, respectively.

The gradient of the reward function is given by:
\begin{equation}
\begin{split}
\nabla_\theta L(\theta) = {} & -\mathbb{E}_{\zeta_i \in D_{demo}} \nabla_\theta r_\theta(\zeta_i) + {}\\
& \textstyle \quad \frac{1}{Z}  \sum_{\zeta_j \in D_{samp}} w_j \nabla_\theta r_\theta(\zeta_j),
\end{split}
\label{eq:reward-grad}
\end{equation}
where $w_j = \frac{\exp(r_\theta(\zeta_j))}{q(\zeta_j)}$ and $Z = \sum_j w_j$.

As shown in Fig.~\ref{fig:reward_model}, our reward model in DG-AIRL consists of two RNN encoders and one MLP network. 
The input encoder compresses the utterances from the context into a context representation which becomes the initial state in the next step. 
The state encoder takes as input the dialogue context and generated words before time $t$ and outputs the new state representation $s_t$ for time step $t$. 
Then the state and action representations are fed to two separate MLP layers respectively. 
The outputs of these two models are concatenated and form the input to the third MLP layer to get the final reward value of current state-action pair $\langle s_t, a_t\rangle $.
\section{Experimental Setup}
\label{sec:experiments}

\subsection{Dataset}
The MovieTriples dataset \citep{serban2016hred} has been developed by expanding and preprocessing the Movie-Dic corpus \citep{banchs2012movie} of film transcripts and each dialogue consists of 3 turns between two interlocutors. 
The dialogues are collected from the scripts of more than 600 movies, which span a wide range of topics. 
We limit the length of the utterances from one speaker in each dialogue turn between $4$ and $80$. 
In the final dataset, there are around 157,000 dialogues in the training set, 19,000 in the validation set, and 19,000 in the test set. 
The average length of each dialogue is about 54. 

\subsection{Experimental settings}
We limit the vocabulary table size to the top 20k most frequent words for the MovieTriples dataset.
All words that are not in the vocabulary tables are replaced with the token `$\langle unk\rangle$'.  Following the preprocessing method from \citep{serban2016hred}, all names and numbers are replaced with the `$\langle person\rangle$ and `$\langle number\rangle$' tokens, respectively \citep{ritter2010unsupervised}. 
Since the context input in each dialogue is made up of several utterances from two different speakers, to capture the interactive structure, we insert a special token `$\langle/s\rangle$' between the first turn and the second. The word embedding size is 200. 

Next, we list the models we consider.
We implement all models based on Tensorflow\footnote{https://www.tensorflow.org/} except VHRED. 

\smallskip\noindent%
\textbf{DG-AIRL.} This is our main model that adopts adversarial inverse reinforcement learning techniques to train a dialogue system. 
The encoder and decoder in the generator (policy network) are built from a 2-layer GRU with 1024 hidden units and an attention mechanism is incorporated into the decoding step. 
With respect to the reward function structure, we choose a 2-layer GRU with 1024 hidden units as the context encoding layer to compress the input to an intermediate representation. 
Then a 1-layer GRU with 1024 hidden units is used to take state-action pair as input and output the next state as shown in Fig.\ref{fig:reward_model}.

\smallskip\noindent%
\textbf{Seq2Seq.} The encoder and decoder in this baseline are copied from the generator in the DG-AIRL model and built from a 2-layer GRU with 1024 hidden units; an attention mechanism is incorporated into the decoding step.

\smallskip\noindent%
\textbf{SeqGan.} This is the model from \citep{li2017adversarial}. In terms of the generator, SeqGan shares the same architecture as the DG-AIRL model.  
With respect to the discriminator in this model, both the first encoder layer and the second context layer are built from a 2-layer GRU with 1024 units separately.    

\smallskip\noindent%
\textbf{VHRED.} For the VHRED model, we reuse the original implementation from the authors, including their tuning techniques.\footnote{For more details, see \url{https://github.com/julianser/hed-dlg-truncated}.}

\smallskip\noindent%
\textbf{DG-AIL.} This is the model with the adversarial imitation learning method, which is also an extension of SeqGan. 
The DG-AIL model shares the same structure as the SeqGan model, including generator and discriminator. The only difference is the loss function, as we discussed in Section~\ref{sec:method}.   

\smallskip\noindent%
We optimize the models using Adam \citep{kingma2014adam} and the learning rate is initialized as 0.001 except for VHRED. 
Dropout with probability 0.3 was applied to the GRUs and we apply gradient clipping for both policy models and reward models. 
We set the beam size to 8 for Monte Carlo search during training and beam search during testing. 
During the training of SeqGan, DG-AIL, and DG-AIRL, we employ the teacher-forcing technique from \citet{li2017adversarial} to increase training efficiency.\footnote{The source code of this work is available at https://bitbucket.org/ZimingLi/dg-irl-aaai2019}

\subsection{Evaluation metrics}
\label{subsection:embedding-metrics}
To evaluate the response quality in dialogue generation, recent work has adopted word-overlap metrics from machine translation to compare a machine-generated response to a single target response.  
Since the response to the input context in dialogue could be very diverse and open, a single target response is not able to cover all reasonable answers. 
\citet{liu2016hownot} show that word-overlap metrics such as BLEU correlate very weakly with reply quality judgments from human annotators. 
To assess the performance of our proposed algorithm, we use two evaluation methods, one is to use word embedding based metrics and the other is to employ human annotators to judge the response quality. We also tried to evaluate the response diversity with the metric \emph{Distinct} but we found that the result is not aligned with the results based on human evaluations; we do not include results based on \emph{Distinct} in this paper.

\subsubsection{Embedding metrics.} With respect to word embedding based methods, we use three  metrics that are also used in \citep{serban2017vhred}:
\begin{itemize}[nosep]
\item \textit{Average embedding}: This method applies cosine-similarity to measure the similarity between the mean word embeddings of the target utterance and the predicted utterance. 
\item \textit{Greedy embedding}: This metric relies on cosine-similarity but adopts greedy matching to find the closest word in the target response for each word in the generated response \citep{rus2012comparison}.
\item \textit{Extrema embedding}: This method computes the word embedding extrema scores \citep{forgues2014bootstrapping} that embed the responses by taking the extrema (maximum of the absolute value) of each dimension, and afterward computes the cosine similarity between them. 
\end{itemize}
A higher score indicates that the generated reply shares similar semantic content with the target response. 
For all three metrics, we use pre-trained Word2Vec word embeddings trained on the Google News Corpus, which is public access.

\subsubsection{Human evaluation.} 
A proper quality evaluation of dialogue responses should cover not only topic-similarity but also lexical aspects, informativeness, interestingness and so on. 
There is currently no reliable metric to assess the overall quality of dialogue responses. 
For this reason, we create human annotations to evaluate the quality of responses given dialogue context with a crowdsourcing platform.\footnote{FigureEight, \url{https://www.figure-eight.com/}.}
Previous work involving human evaluation usually has two experimental settings: pairwise comparison and pointwise scoring. 
In pairwise comparisons, annotators are asked to choose the better response from replies generated by two models while in the pointwise method, annotators are asked to rate the overall quality of each response, typically on a scale from $0$ (low quality) to $4$ (high quality). 
We employ both pairwise and pointwise assessments.
We use a pairwise setting to directly contrast the overall performance of our model against others. 
Pointwise scoring may be noisier than pairwise judgments since human annotators need to give an exact score.
However, pointwise judgments give us a chance to analyze the differences between replies at a fine-grained level of detail. 

We randomly sample 1,000 dialogue contexts from the test set of the MovieTriple dataset. 
Each context has five replies from five generation models and we have 5,000 context-reply pairs in total; 2,500 are used for pointwise scoring while the remaining 2,500 are grouped into 2,000 comparison pairs for the pairwise setting. 
Each comparison pair has one dialogue context and two replies, where one is from our DG-AIRL model and the other is from a baseline model.

For the pairwise setting, we ask annotators to judge which of two responses is more appropriate given a dialogue context. 
We instruct annotators which aspects they should take into account when making a decision. 
The top priority is that an appropriate response must be relevant; in addition, they should consider:
\begin{itemize}[nosep]
\item whether the response is natural;
\item whether the response is interesting;
\item whether the response can make the conversation continue, that is, whether the response is proactive; and
\item whether the response is the only possible reply to the given context.
\end{itemize}
If the annotators think neither of the responses is more appropriate or it is impossible to infer the conversation from the given context, they are asked to choose the third choice -- ``Neither is more appropriate." 
We insert test questions to exclude annotators who lack the capacity to finish the tasks, such as limited English skills. 
We only accept annotators considered ``highly trusted'' by the crowdsourcing platform and require $90\%$ accuracy on designed ``test questions.'' 
Each comparison pair is assessed by three annotators.

For pointwise judgments, annotators were asked to judge the overall quality from $0$ to $2$:
\begin{description}[nosep]
\item[\rm +2:] The response is not only relevant and natural, but also informative and interesting; 
the response need not be so interesting, but it is natural and can make the conversation continue (more proactive); the response is the only possible reply to the dialogue.
\item[\rm +1:] The response can be used as a reply to the context, but it is too generic like ``I don't know.'' These replies are usually more reactive. 
\item[\rm 0:] The response cannot be used as a reply to the context. 
It is either semantically irrelevant or disfluent.
\end{description}
At the start of the annotation effort, we instruct the annotators and show them several examples of how to assign grades to a given dialogue. 
We use the same quality checks and annotator selection criteria as in the pairwise setting.
Each context-reply is assigned to three human annotators.

\section{Results and Analysis}
\label{sec:results}
\subsection{Results using embedding metrics}
In Table~\ref{Table:embed_table}, we report the scores obtained using the embedding metrics (Section~\ref{subsection:embedding-metrics}). 
All response generation models are fine-tuned to obtain the highest score on the validation dataset. 
We found $0.01$ and $0.1$ to be the optimal values of $\lambda$ for the DG-AIL model and the DG-AIRL model. 

The DG-AIRL and DG-AIL models achieve the highest scores using the embedding metrics, which means that they can better capture the topic of the target response than other models. 
\begin{table}[h]
  \centering
  \resizebox{\columnwidth}{!}{
  \begin{tabular}{ lc*{5}{c}}
    \toprule
\textbf{Model}  &\makecell{\textbf{Average}} & \makecell{\textbf{Greedy}} & \makecell{\textbf{Extrema}}  & \makecell{\textbf{Length}}\\
\midrule
Seq2Seq     & $0.563 \pm 0.003$    & $0.167 \pm 0.001$   & $0.352 \pm 0.002$   & 8.8 \\
SeqGan        & $0.564 \pm 0.003$    & $0.165 \pm 0.001$   & $0.354 \pm 0.002$   & 9.7 \\
VHRED        & $0.507 \pm 0.003$    & $0.145 \pm 0.001$   & $0.309 \pm 0.002$   & \textbf{12.0}  \\
DG-AIL        & $0.553 \pm 0.003$    & \boldmath $0.171^* \pm 0.001$   & $0.356 \pm 0.002$    & 7.7\\
DG-AIRL       &\boldmath $0.589^* \pm 0.003$ & $0.169 \pm 0.001$   & \boldmath $0.368^* \pm 0.002$ & 10     \\
    \bottomrule
  \end{tabular}
  }
\vspace*{-.5\baselineskip}
\caption{Performance in terms of  embedding metrics of response generation models, with $95\%$ confidence intervals. * indicates the result is statistically significant ($p < 0.005$) with a paired t-test over DG-AIRL and other baseline models.} 
\label{Table:embed_table}
\end{table}
The performance of the VHRED model is unexpected since this method achieves the lowest value while it is one of the state-of-the-art methods in dialogue response generation. 
A possible reason is that the other four models adopt an attention mechanism to directly capture the relation between generated words and input words from context. 
\citet{serban2017vhred} state that VHRED produces longer responses and its responses are on average more diverse based on unigram entropy. 
Besides, DG-AIL and DG-AIRL are also supposed to generate more diverse responses.
However, a more diverse response does not mean it is an appropriate response. If the response deviates too much from the target topic, the response content will not be relevant to the dialogue context and it deserves a lower quality score. On the other hand, if a diverse response is appropriate to the dialogue context, it is unfair to use embedding-based metrics to assess these kinds of generative models. 

Given these considerations, we believe that embedding-based metrics may not be powerful enough to reflect the overall response quality and that it is essential to carry out human evaluations. 

\subsection{Results using human annotations}
\subsubsection{Pairwise evaluation.} 
As shown in Table \ref{Table:pairwise_table}, our model DG-AIRL outperforms other response generation models based on pairwise comparisons. 
Among the first three models, the DG-AIRL model outperforms them all at the probability 0.46. 
\begin{table}[t]
  \centering
\resizebox{.6\linewidth}{!}{
\begin{tabular}{l*{3}{c}}
\toprule

\textbf{Model pair} & \textbf{Win} & \textbf{Tie} & \textbf{Loss} \\
\midrule
DG-AIRL-Seq2Seq   & \textbf{0.44}     & 0.29  & 0.27  \\
DG-AIRL-VHRED  & \textbf{0.46}   & 0.32 & 0.22  \\
DG-AIRL-SeqGan   & \textbf{0.47} & 0.25 & 0.28     \\
DG-AIRL-DG-AIL  & 0.36 & \textbf{0.37} & 0.27  \\
\bottomrule
\end{tabular}
}
\vspace*{-.5\baselineskip}
\caption{Performance in terms of pairwise human annotations of response generation models.} 
\label{Table:pairwise_table}
\end{table}
\begin{table}[t]
  \centering
\resizebox{.9\columnwidth}{!}{
\begin{tabular}{l*{4}{c}}
\toprule
\textbf{Model} & \textbf{Freq of +2} & \textbf{Freq of +1} & \textbf{Freq of 0} & \textbf{Avg Score}\\
\midrule
Seq2Seq   & 0.09  & 0.22  & 0.69   & 0.40\\
SeqGan    & 0.09  & 0.21  & 0.70   & 0.39\\
VHRED     & 0.12  & 0.25  & 0.63   & 0.49\\
DG-AIL       & 0.12  & 0.29  & 0.59   & 0.53\\
DG-AIRL       & 0.13  & 0.28  & 0.59   & \textbf{0.54}\\
\bottomrule
\end{tabular}
}
\vspace*{-.5\baselineskip}
\caption{Performance in terms of pointwise human evaluations of response generation models. ``Freq of $N$'' is the relative frequency of a model's responses with a score of $N$.} 
\vspace*{-1\baselineskip}
\label{Table:pointwise_table}
\end{table}
The difference is that the \textbf{Win} rate of VHRED (the lose rate of DG-AIRL) is lower than Seq2Seq and SeqGan. 
In other words, although VHRED has a higher probability to be tied with DG-AIRL, it loses more compared to Seq2Seq and SeqGan. 

As we said in the last section, a possible explanation for these results is that VHRED does produce longer responses (Table \ref{Table:embed_table}) but the contents of these responses deviate too much from the target topic, which could result in lower performance. 
In our human evaluation setting, fluency is not the only aspect annotators need to consider while determining their preference for a response. 
By taking into account different factors, such as relevance, fluency, informativeness, we think the final judgments from human annotators are trustworthy. 
The Fleiss' kappa score, which indicates the agreement among labelers \citep{fleiss1973equivalence}, is around 0.23. 
This value is not high and a possible reason is that judging the response quality is challenging for human annotators when only 1 or 2 utterances are provided as context, especially on the MovieTriples dataset.

\begin{table*}[!t]
  \centering
\resizebox{\linewidth}{!}{
  \begin{tabular}{ c  c }
    \toprule
    \textbf{Context} & \textbf{Next Reply from Speaker A} \\
    \midrule
    \makecell[l]{\textbf{Speaker A:} we did a story on this guy . he owns half of arizona .\\ 
                  \textbf{Speaker B:} is he a fraud ?}  
    & \makecell[l]{\textbf{Reference:} how would you ever know ?\\ 
                 \textbf{Seq2Seq:} yeah . no . i do not know who he is .\\
                \textbf{SeqGan:} sure , i guess not . \\ 
                \textbf{VHRED:} he doesn't want to know . \\ 
                \textbf{DG-AIL:} no . no , he is not .  \\
                \textbf{DG-AIRL:} no . maybe he is a good man .} \\
    \midrule

    \makecell[l]{\textbf{Speaker A:} and i am honored that you want to be with me because \\ you would never be with anyone who was not truly worthy . \\ 
                 \textbf{Speaker B:} i feel exactly the same way about you .}  
    & \makecell[l]{\textbf{Reference:}   oh, $\langle person \rangle$, do not say that , please , that just makes it worse .\\
                  \textbf{Seq2Seq:}  do you love your $\langle unk \rangle$ .\\
                \textbf{SeqGan:}  can we i . you . you . you . you . you . you . . you . \\
                \textbf{VHRED:}   $\langle person \rangle$ ?\\ 
                \textbf{DG-AIL:}   what is it .\\
                 \textbf{DG-AIRL:}  is that what you think .} \\
    \midrule

    \makecell[l]{\textbf{Speaker A:}  oh , that was not my father .\\
                   \textbf{Speaker B:}  what ?}  
    & \makecell[l]{\textbf{Reference:} that was not him . i heard him tell the cops .\\ 
                   \textbf{Seq2Seq:} i am sorry . i needed to know how i felt about it .\\
                \textbf{SeqGan:} you said he was a singer . \\ 
                 \textbf{VHRED:} i can't . i'm sorry . \\ 
                \textbf{DG-AIL:} i do not know . i guess . \\
                 \textbf{DG-AIRL:} it is not him . i do not know what happened .} \\
    \bottomrule
  \end{tabular}
}
\vspace*{-.5\baselineskip}
\caption{Example output. People names are replaced with the token `$\langle person\rangle$'. All words are converted to lowercase.} 
\vspace*{-1\baselineskip}
\label{Table:case_table}
\end{table*}
Compared to the first three models, the DG-AIL model achieves a better performance. 
According to the performance of DG-AIL and SeqGan, we can claim that causal entropy regularization improves the performance of dialogue models that employ adversarial training because they share the same structure and the only difference is the entropy regularization term. 
This also echoes the usefulness of entropy regularization in adversarial models. 
Although the DG-AIRL model still beats DG-AIL, the difference between the \textbf{Win} and the \textbf{Loss} rates is much smaller compared to other models. 
DG-AIL model has a higher chance to draw with the DG-AIRL model. 
These two models both adopt entropy regularization and they have the same generator structure. 
The performance difference comes from the reward model: DG-AIRL is able to improve the response quality because it has a specific reward model for each state-action pair and adopts importance sampling. 
The reward signal in DG-AIRL is more concrete and reliable compared to DG-AIL.

\subsubsection{Pointwise evaluation.} 
The evaluation results based on pointwise judgments are shown in Table \ref{Table:pointwise_table}. 
According to the average score, the DG-AIRL and DG-AIL models outperform other models and the performance of DG-AIRL and DG-AIL are quite close. 
This does not mean the result of the pointwise evaluation is in conflict with the conclusion we made in the last section, viz.\ that the DG-AIRL model beats the DG-AIL model based on the pairwise evaluation. 
Compared to the pairwise evaluation, the pointwise evaluation needs to assign an exact score to each context-reply pair and this score is independent of the other replies to the same dialogue context. 
In contrast,  in the pairwise evaluation we consider a pair of replies to the same context at the same time and it is more natural and reliable if we want a ranked list based on performance. 
The advantage of the pointwise setting is that it can provide quality distributions of different models and help us find out what makes a model performance different.

As shown in Table \ref{Table:pointwise_table}, we find that VHRED, DG-AIL, and DG-AIRL generate almost the same number of high-quality responses (responses that received a score ``+2"). 
The VHRED model loses the competition with DG-AIRL and DG-AIL models because it generates more low-quality replies (responses that get score ``0"). 
In our experimental setup, we ask annotators to grade the reply quality as ``+1" (fine quality) if the response can be used as a reply to the message, but is too generic. 
In Section~\ref{sec:introduction}, we expect to generate more diverse responses and avoid producing too generic responses, such as ``I don't know". 
However, in some dialogue contexts, ``I don't know" is still an appropriate and reasonable response. 
As shown in Table \ref{Table:case_table}, DG-AIRL and DG-AIL improve the proportion of high-quality responses without losing the capacity to generate fine quality replies. 

\section{Related Work}
Based on developments in sequential neural networks,  \citet{shang2015neural} and \citet{sordoni2015seq2seq} propose to generate high-quality replies in a dialogue system with a recurrent neural network. 
To formulate the complex dependencies between different utterances in multi-turn dialogs, \citet{serban2016hred} propose to adopt a hierarchical recurrent encoder-decoder neural network (HRED) to the dialogue domain, where word-level and utterance-level \ac{RNN} are used. 

Built on HRED, the same group of authors create a more powerful generative architecture \citep{serban2017vhred} with latent stochastic variables that span a variable number of time steps (VHRED). 
To train these \ac{RNN} models, supervised training is commonly used, which minimize the cross-entropy between the generated reply and an oracle reply. 
However, in terms of open-domain dialogue systems, there could be multiple reasonable replies for the same input context. 
In other words, the entropy of the target replies is high. 

\citet{li2017adversarial} cast the task of open-domain dialogue generation as an \ac{RL} problem and train a generator based on the signal from a discriminator to generate response sequences indistinguishable from human-generated dialogs.

\section{Conclusion}
In this work, we have investigated two adversarial training methods for open-domain dialogue systems. 
We have first adopted adversarial imitation learning to force our model to generate human-like dialogue responses. 
Besides that, we have incorporated an entropy regularization term to the generator objective function, which can alleviate the problem of model collapse. 
Our second and main method, DG-AIRL, relies on techniques of adversarial inverse reinforcement learning. 
We design a specific reward architecture to supply a more accurate and precise reward signal for the generator training. 

To assess the overall performance of our models, we propose two human-evaluation settings. 
We adopt the results from a pairwise evaluation setting to show that our model can outperform state-of-the-art methods in open-domain dialogue generation. 
To analyze the different in replies from different models, we explore the results from a pointwise evaluation setting, which can provide a general quality distribution for different models.

In future work, we plan to extend the idea of reward learning to multi-turn dialogue generation, which can propagate the reward signal between conversation turns. 
Another promising research direction is to explore the usefulness of recovered reward models, for instance to evaluate the quality of generated responses from other models.

\section*{Acknowledgements}
This research was partially supported by 
the China Scholarship Council
and
the Google Faculty Research Awards program.
All content represents the opinion of the authors, which is not necessarily shared or endorsed by their respective employers and/or sponsors.

\nocite{*}
\bibliographystyle{aaai} 
\bibliography{bibliography}

\begin{thebibliography}{}

\bibitem[\protect\citeauthoryear{Bahdanau, Cho, and
  Bengio}{2014}]{bahdanau2014attention}
Bahdanau, D.; Cho, K.; and Bengio, Y.
\newblock 2014.
\newblock Neural machine translation by jointly learning to align and
  translate.
\newblock {\em arXiv preprint arXiv:1409.0473}.

\bibitem[\protect\citeauthoryear{Banchs}{2012}]{banchs2012movie}
Banchs, R.~E.
\newblock 2012.
\newblock Movie-dic: a movie dialogue corpus for research and development.
\newblock In {\em ACL: Short Papers-Volume 2},  203--207.
\newblock ACL.

\bibitem[\protect\citeauthoryear{Finn \bgroup et al\mbox.\egroup
  }{2016}]{finn2016connection}
Finn, C.; Christiano, P.; Abbeel, P.; and Levine, S.
\newblock 2016.
\newblock A connection between generative adversarial networks, inverse
  reinforcement learning, and energy-based models.
\newblock {\em arXiv preprint arXiv:1611.03852}.

\bibitem[\protect\citeauthoryear{Finn, Levine, and
  Abbeel}{2016}]{guided_cost_learning}
Finn, C.; Levine, S.; and Abbeel, P.
\newblock 2016.
\newblock Guided cost learning: Deep inverse optimal control via policy
  optimization.
\newblock In {\em ICML},  49--58.

\bibitem[\protect\citeauthoryear{Fleiss and
  Cohen}{1973}]{fleiss1973equivalence}
Fleiss, J.~L., and Cohen, J.
\newblock 1973.
\newblock The equivalence of weighted kappa and the intraclass correlation
  coefficient as measures of reliability.
\newblock {\em Educational and Psychological Measurement} 33(3):613--619.

\bibitem[\protect\citeauthoryear{Forgues \bgroup et al\mbox.\egroup
  }{2014}]{forgues2014bootstrapping}
Forgues, G.; Pineau, J.; Larchev{\^e}que, J.-M.; and Tremblay, R.
\newblock 2014.
\newblock Bootstrapping dialog systems with word embeddings.
\newblock In {\em NIPS, Modern Machine Learning and Natural Language Processing
  Workshop}, volume~2.

\bibitem[\protect\citeauthoryear{Fu, Luo, and Levine}{2017}]{fu2017learning}
Fu, J.; Luo, K.; and Levine, S.
\newblock 2017.
\newblock Learning robust rewards with adversarial inverse reinforcement
  learning.
\newblock {\em arXiv preprint arXiv:1710.11248}.

\bibitem[\protect\citeauthoryear{Ho and Ermon}{2016}]{gan_imitation}
Ho, J., and Ermon, S.
\newblock 2016.
\newblock Generative adversarial imitation learning.
\newblock In {\em NIPS},  4565--4573.

\bibitem[\protect\citeauthoryear{Kingma and Ba}{2014}]{kingma2014adam}
Kingma, D.~P., and Ba, J.
\newblock 2014.
\newblock Adam: A method for stochastic optimization.
\newblock {\em arXiv preprint arXiv:1412.6980}.

\bibitem[\protect\citeauthoryear{Kramer}{1998}]{kramer1998directed}
Kramer, G.
\newblock 1998.
\newblock {\em Directed information for channels with feedback}.
\newblock Ph.D. Dissertation, Swiss Federal Institute of Technology Zurich.

\bibitem[\protect\citeauthoryear{Li \bgroup et al\mbox.\egroup
  }{2016a}]{li2016diversity}
Li, J.; Galley, M.; Brockett, C.; Gao, J.; and Dolan, B.
\newblock 2016a.
\newblock A diversity-promoting objective function for neural conversation
  models.
\newblock In {\em NAACL},  110--119.

\bibitem[\protect\citeauthoryear{Li \bgroup et al\mbox.\egroup
  }{2016b}]{li2016deeprl}
Li, J.; Monroe, W.; Ritter, A.; Jurafsky, D.; Galley, M.; and Gao, J.
\newblock 2016b.
\newblock Deep reinforcement learning for dialogue generation.
\newblock In {\em EMNLP},  1192--1202.

\bibitem[\protect\citeauthoryear{Li \bgroup et al\mbox.\egroup
  }{2017}]{li2017adversarial}
Li, J.; Monroe, W.; Shi, T.; Jean, S.; Ritter, A.; and Jurafsky, D.
\newblock 2017.
\newblock Adversarial learning for neural dialogue generation.
\newblock In {\em EMNLP},  2157--2169.

\bibitem[\protect\citeauthoryear{Liu \bgroup et al\mbox.\egroup
  }{2016}]{liu2016hownot}
Liu, C.-W.; Lowe, R.; Serban, I.; Noseworthy, M.; Charlin, L.; and Pineau, J.
\newblock 2016.
\newblock How not to evaluate your dialogue system: An empirical study of
  unsupervised evaluation metrics for dialogue response generation.
\newblock In {\em EMNLP},  2122--2132.

\bibitem[\protect\citeauthoryear{Permuter, Kim, and
  Weissman}{2008}]{permuter2008directed}
Permuter, H.~H.; Kim, Y.-H.; and Weissman, T.
\newblock 2008.
\newblock On directed information and gambling.
\newblock In {\em Information Theory, 2008. ISIT 2008. IEEE International
  Symposium on},  1403--1407.
\newblock IEEE.

\bibitem[\protect\citeauthoryear{Ritter, Cherry, and
  Dolan}{2010}]{ritter2010unsupervised}
Ritter, A.; Cherry, C.; and Dolan, B.
\newblock 2010.
\newblock Unsupervised modeling of twitter conversations.
\newblock In {\em NAACL},  172--180.
\newblock Association for Computational Linguistics.

\bibitem[\protect\citeauthoryear{Ritter, Cherry, and
  Dolan}{2011}]{ritter2011data}
Ritter, A.; Cherry, C.; and Dolan, W.~B.
\newblock 2011.
\newblock Data-driven response generation in social media.
\newblock In {\em EMNLP},  583--593.
\newblock Association for Computational Linguistics.

\bibitem[\protect\citeauthoryear{Rus and Lintean}{2012}]{rus2012comparison}
Rus, V., and Lintean, M.
\newblock 2012.
\newblock A comparison of greedy and optimal assessment of natural language
  student input using word-to-word similarity metrics.
\newblock In {\em Proceedings of the Seventh Workshop on Building Educational
  Applications Using NLP},  157--162.
\newblock ACL.

\bibitem[\protect\citeauthoryear{Serban \bgroup et al\mbox.\egroup
  }{2016}]{serban2016hred}
Serban, I.~V.; Sordoni, A.; Bengio, Y.; Courville, A.~C.; and Pineau, J.
\newblock 2016.
\newblock Building end-to-end dialogue systems using generative hierarchical
  neural network models.
\newblock In {\em AAAI}, volume~16,  3776--3784.

\bibitem[\protect\citeauthoryear{Serban \bgroup et al\mbox.\egroup
  }{2017}]{serban2017vhred}
Serban, I.~V.; Sordoni, A.; Lowe, R.; Charlin, L.; Pineau, J.; Courville,
  A.~C.; and Bengio, Y.
\newblock 2017.
\newblock A hierarchical latent variable encoder-decoder model for generating
  dialogues.
\newblock In {\em AAAI},  3295--3301.

\bibitem[\protect\citeauthoryear{Shang, Lu, and Li}{2015}]{shang2015neural}
Shang, L.; Lu, Z.; and Li, H.
\newblock 2015.
\newblock Neural responding machine for short-text conversation.
\newblock In {\em ACL}, volume~1,  1577--1586.

\bibitem[\protect\citeauthoryear{Shi \bgroup et al\mbox.\egroup
  }{2018}]{shi2018towards}
Shi, Z.; Chen, X.; Qiu, X.; and Huang, X.
\newblock 2018.
\newblock Towards diverse text generation with inverse reinforcement learning.
\newblock {\em arXiv preprint arXiv:1804.11258}.

\bibitem[\protect\citeauthoryear{Sordoni \bgroup et al\mbox.\egroup
  }{2015a}]{sordoni2015neural}
Sordoni, A.; Galley, M.; Auli, M.; Brockett, C.; Ji, Y.; Mitchell, M.; Nie,
  J.-Y.; Gao, J.; and Dolan, B.
\newblock 2015a.
\newblock A neural network approach to context-sensitive generation of
  conversational responses.
\newblock In {\em NAACL},  196--205.

\bibitem[\protect\citeauthoryear{Sordoni \bgroup et al\mbox.\egroup
  }{2015b}]{sordoni2015seq2seq}
Sordoni, A.; Galley, M.; Auli, M.; Brockett, C.; Ji, Y.; Mitchell, M.; Nie,
  J.-Y.; Gao, J.; and Dolan, B.
\newblock 2015b.
\newblock A neural network approach to context-sensitive generation of
  conversational responses.
\newblock {\em arXiv preprint arXiv:1506.06714}.

\bibitem[\protect\citeauthoryear{Wen \bgroup et al\mbox.\egroup
  }{2016}]{wen2016network}
Wen, T.-H.; Vandyke, D.; Mrksic, N.; Gasic, M.; Rojas-Barahona, L.~M.; Su,
  P.-H.; Ultes, S.; and Young, S.
\newblock 2016.
\newblock A network-based end-to-end trainable task-oriented dialogue system.
\newblock {\em arXiv preprint arXiv:1604.04562}.

\bibitem[\protect\citeauthoryear{Xing \bgroup et al\mbox.\egroup
  }{2017}]{xing2017topic}
Xing, C.; Wu, W.; Wu, Y.; Liu, J.; Huang, Y.; Zhou, M.; and Ma, W.-Y.
\newblock 2017.
\newblock Topic aware neural response generation.
\newblock In {\em AAAI}, volume~17,  3351--3357.

\bibitem[\protect\citeauthoryear{Yu \bgroup et al\mbox.\egroup
  }{2016}]{yu2016strategy}
Yu, Z.; Xu, Z.; Black, A.~W.; and Rudnicky, A.
\newblock 2016.
\newblock Strategy and policy learning for non-task-oriented conversational
  systems.
\newblock In {\em SIGDIAL},  404--412.

\bibitem[\protect\citeauthoryear{Ziebart, Bagnell, and
  Dey}{2010}]{ziebart2010causal-entropy}
Ziebart, B.~D.; Bagnell, J.~A.; and Dey, A.~K.
\newblock 2010.
\newblock Modeling interaction via the principle of maximum causal entropy.
\newblock In {\em ICML}.

\bibitem[\protect\citeauthoryear{Ziebart \bgroup et al\mbox.\egroup
  }{2008}]{ziebart_aaai_2008}
Ziebart, B.~D.; Maas, A.~L.; Bagnell, J.~A.; and Dey, A.~K.
\newblock 2008.
\newblock Maximum entropy inverse reinforcement learning.
\newblock In {\em AAAI},  1433--1438.
\newblock {AAAI} Press.

\end{thebibliography}

\end{document}